\documentclass[lettersize,journal]{IEEEtran}
\usepackage[utf8]{inputenc}
\usepackage{times}
\usepackage{epsfig}
\usepackage{subfigure}
\usepackage{graphicx}
\usepackage{amsmath}
\usepackage{amssymb}
\usepackage{authblk}
\usepackage[ruled,linesnumbered]{algorithm2e}
\usepackage[pagebackref=true,breaklinks=true,colorlinks,bookmarks=false]{hyperref}

\begin{document}

\title{Neural Network Compression via Effective Filter Analysis and Hierarchical Pruning}







\author{Ziqi Zhou, Li Lian, Yilong Yin*, Ze Wang*~\IEEEmembership{Senior Member,~IEEE,}%
\IEEEcompsocitemizethanks{
\IEEEcompsocthanksitem This research was funded by the National Natural Science
Foundation of China (61876098, 61573219) (YY) and by NIH Grants: R01AG060054, R01 AG070227, R01EB031080, P41EB029460 (ZW).
\IEEEcompsocthanksitem Ze Wang is with the Department of Diagnostic Radiology and Nuclear Medicine, University of Maryland and School of Medicine, Baltimore, MD, 30332.
\IEEEcompsocthanksitem Yilong Yin is with Research Center of Artificial Intelligence, Shandong University and School of Software, Jinan, China, 250101.
\IEEEcompsocthanksitem Li Lian is with Research Center of Artificial Intelligence, Shandong University and School of Software, Jinan, China, 250101.
\IEEEcompsocthanksitem Ziqi Zhou is with Research Center of Artificial Intelligence, Shandong University and School of Software, Jinan, China, 250101.
\IEEEcompsocthanksitem Corresponding Authors: Ze Wang(ze.wang@som.umaryland.edu) and Yilong Yin(ylyin@sdu.edu.cn)
}%
}




\maketitle

\begin{abstract}
Network compression is crucial to making the deep networks to be more efficient, faster, and generalizable to low-end hardware. Current network compression methods have two open problems: first, there lacks a theoretical framework to estimate the maximum compression rate; second, some layers may get over-prunned, resulting in significant network performance drop. To solve these two problems, this study propose a gradient-matrix singularity analysis-based method to estimate the maximum network redundancy. Guided by that maximum rate, a novel and efficient hierarchical network pruning algorithm is developed to maximally condense the neuronal network structure without sacrificing network performance. Substantial experiments are performed to demonstrate the efficacy of the new method for pruning several advanced convolutional neural network (CNN) architectures. Compared to existing pruning methods, the proposed pruning algorithm achieved state-of-the-art performance. At the same or similar compression ratio, the new method provided the highest network prediction accuracy as compared to other methods.
\end{abstract}




\section{Introduction}

\begin{figure*}[tbp]
\begin{center}
    \includegraphics[width=0.8\linewidth]{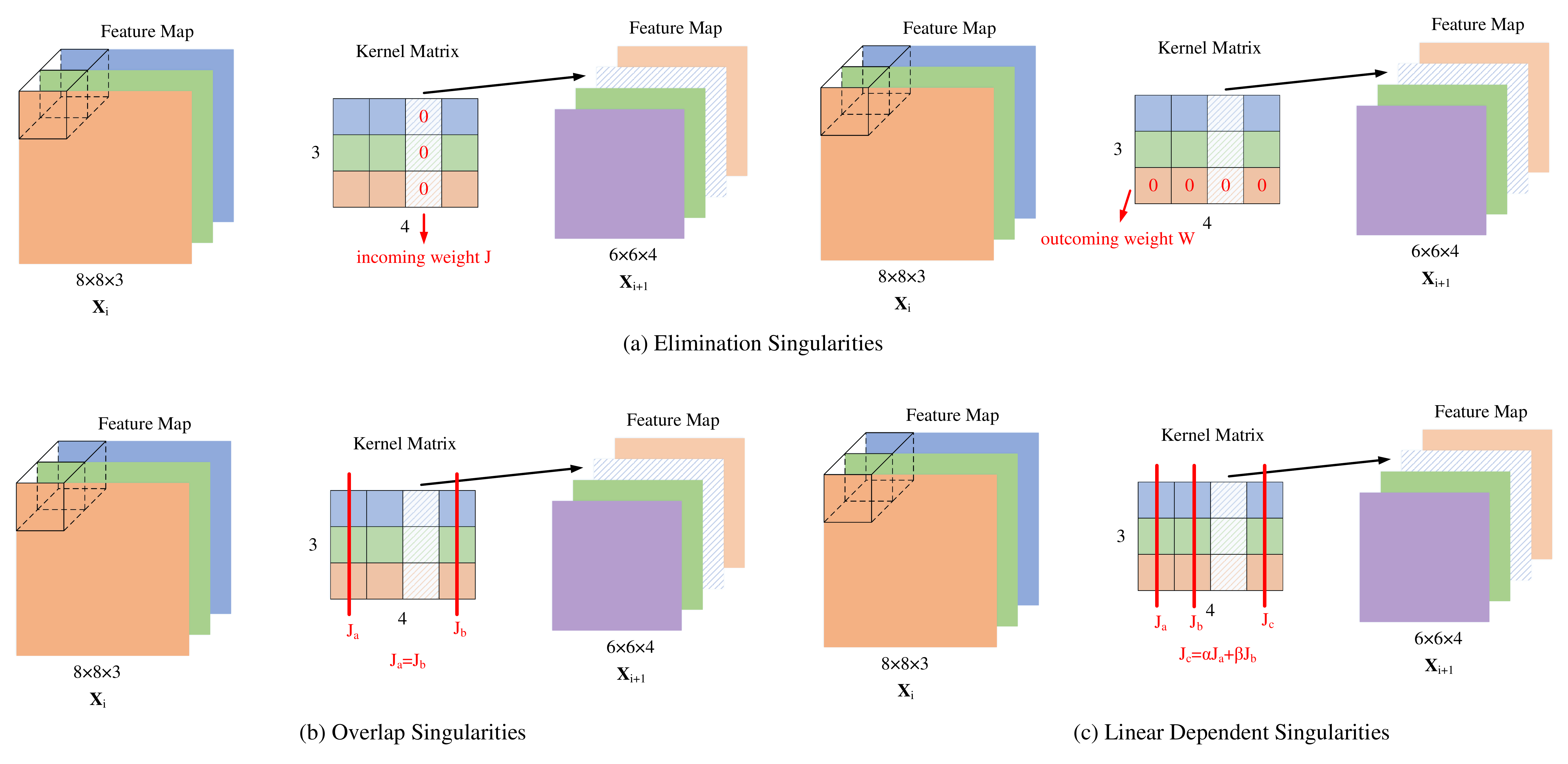}
\end{center}
    \caption{Three types of singularities in neural networks as described in \cite{orhan2018skip}. Overlap singularity is induced by the permutation symmetry of hidden nodes. This type of singularity generates identical loss function derivatives with respect to two distinct hidden units in the same layer for the same inputs. The elimination singularity is caused by the zero incoming weight or zero input, which actually indicates that the corresponding units are completely ineffective and should be removed. The linearity-induced singularity is caused by the linear dependence of the inputs, making a set of units in the same layer linearly dependent. }
    \label{Singularities}
\end{figure*}
\IEEEPARstart{W}{ith} the help of fast parallel computing, especially through the graphic process unit (GPU) cards, deep neural networks have quickly become the state-of-the-art in many research fields such as image classification, image generation, medical image analysis etc. While deep networks with more layers of neurons with more connections tend to have better performance for many applications such as the well-known ImageNet classification competitions \cite{AlexNet2012NIPS, VggNet2015ICLR,GoogleNet2015CVPR,ResNet2016CVPR}, recent research also showed that further increasing network size after the network complexity reaches a certain level only produced very little performance gain \cite{montavon2010layer,Zhang2017Overparameterized}. The little performance gain was even at the expense of an escalated risk of model overfitting because bigger networks have more parameters to estimate. The increased computation time of bigger networks is also problematic for applications oriented for mobile or embedded devices. Over the years, various methods have been proposed to minimize the network structure without sacrificing network prediction accuracy. With a pre-specified compression ratio, a widely used network compression (or pruning) approach is to iteratively eliminate the unimportant connections or neurons that show weak influence to the network output. The connection-level pruning  is to make the weight matrices as sparse as possible \cite{Han2015compression, Han2015Efficient, Guo2016Dynamic, Dong2017Surgeon, Perpinan2018LCA, Tung2018CLIP-Q} is to make the weight matrices as sparse as possible. The neuron-wise pruning, also called filter pruning or channel-level pruning, is to identify and remove the redundant or weakly contributing channels \cite{liu2017learning, luo2017thinet, He2017Accelerating, X.Suau2018PCA, Wang2018Linear, yu2018nisp, Zhuang2018Discrimination, Lin2019Generative}. This paper focuses on neuron-wise pruning (or filter pruning) because it is more effective than the connection-wise approach to achieve the overall compression ratio and is more flexible for hardware implementation because it does not need to save the sparse weights.  

Each iteration of the neuron-wise pruning (filter pruning) contains two components: filter salience evaluation and pruning. At each iteration, a salience score is calculated for each filter and compared to a predefined threshold. Those below the threshold will be treated as unimportant channels and be subsequently removed from the network. One problem is that there still lacks a theoretical framework to estimate the maximum compression ratio. The target compression ratio can only be achieved through trials and errors until the pre-specified compression ratio is reached or a significant network performance drop is observed. Filter pruning is often done through a layer-independent filter removal, i.e., filters marked as unimportant will be removed from all layers simultaneously. This approach is efficient but can cause a layer-overpruning problem: some layers may have much more filters removed than others and the extremely inhomogeneous pruning may result in abrupt network performance drop due to the sudden information transition bottleneck in those layers \cite{Han2015Efficient,He2017Accelerating, li2017pruning, luo2017thinet, yu2018nisp}. 
Filter-wise pruning is less problematic for having the layer-overpruning issue but it is extremely slow as there may be tens of thousands of neurons to be assessed. The purpose of this study was to solve these problems using two novel methods. The maximum network redundancy and the associated compression ratio were estimated by a gradient matrix deficit analysis derived from the Hessian matrix degeneration analysis \cite{orhan2018skip}. The layer-overpruning and overall network pruning efficiency problems were addressed using a novel hierarchical layer-wise and then filter-wise pruning algorithm.  

Hessian matrix degeneration analysis was proposed in \cite{orhan2018skip} to explain the working mechanism of skip connections, which is now a standard network structure in deep network. According to Orhan and Pitkow\cite{orhan2018skip}, deep neural networks often have three different types of singularities\ref{Singularities}: the overlap singularity, the elimination singularity, and the linear dependence singularity. All three types of singularities will cause rank deficits for the Hessian matrix and subsequently cause network training uncertainty. As a result, only a portion of network weight changes will have an impact on loss function during training; the rest will have little or no effect. The skip connections can effectively remove the singularities of the Hessian loss function by directly bypassing these singular connections. Reciprocally, the rank of the Hessian matrix can guide how many skip connections a network can have in order to achieve the best network performance. The same idea can be used to estimate network filter redundancy and subsequently the optimal network pruning ratio. Computing the Hessian matrix is highly computationally expensive as it involves all training data. Instead, we trained the model as usual until it reached the optimum. We then calculated the weight gradient matrix to estimate the network redundancy and the optimal pruning ratio. As described later, in theory, the covariance matrix of weight gradient is the scaled version of the Hessian matrix when the network weights reach the optima. We then used Principal Component Analysis (PCA)\cite{Hotellings1933PCA} to estimate the matrix rank. Because network weight training in deep learning is often based on batches of training samples using the stochastic gradient descent method, the stochastic errors will cause an avoidable discrepancy between the learned data distributions and true data distributions, then it occurs a deviation between the learned training direction and the true training direction. These errors will affect the subsequent Hessian matrix analysis. To minimize the stochastic errors induced variations, we introduced a weight gradient matrix pre-processing step. We first used first-order Taylor approximation of the loss function to estimate the contribution of weights to the loss function. For those whose contributions are nearly zero, we set the gradients with respect to them to be 0. We then applied PCA to this preprocessed gradient matrix, determined the number of non-dominant eigenvectors of each layer, and added up the number of dominant eigenvectors of each layer to get the total number of non-dominant eigenvectors of the whole neural network. The ratio of the total number of non-dominant eigenvectors to the total number of feature channels was used as the `filter pruning ratio' for each pruning step of our method. After finding the `filter pruning ratio', cross entropy of the weight matrix of two adjacent layers was calculated to determine which layers should be removed. The next level pruning of our hierarchical method was applied to each filter separately.

Below are the major contributions of this study:
\begin{itemize}
    \item A PCA-based weight gradient matrix deposition method for estimating the optimal pruning ratio. This method is based on the singularity analysis of the Hessian matrix and offers a new theoretical framework for the optimal pruning ratio estimation; 
    \item A weight gradient matrix preprocessing method to reduce the stochastic errors to the loss function and subsequently to the weight gradients;
    \item The connection between the Hessian matrix of loss function and the expectation of the covariance of the gradient;
    \item A hierarchical pruning approach which combines a layer-wise and a subsequent filter-wise pruning. The cross-entropy based least contributing layer determination process is also new.
\end{itemize}

The rest of the paper is organized as follows: related work is provided in Section 2; method details are given in Section 3; experiments and results are detailed in Section 4;  the results and the proposed method are explained and discussed in Section 5.

\section{Related Work}
Based on the pruning criteria, current filter pruning methods can be roughly divided into four categories. 

\textbf{Weight or activation strength-based methods. }This type of method can reduce the eliminating singularity by removing weak connections with negligible weights or activation. To enforce the weight sparsity, additional loss function regularization such as the L1-norm of the weights is often added during network training \cite{li2017pruning}. The same idea was used to remove filters or layers by enhancing the filter or layer-wise sparsity using group Lasso-based loss function regularization \cite{wen2016learning} or L1 regularization on the batch normalization layer \cite{liu2017learning}. Rather than using the weight strength as the pruning criterion, activation can be used to identify candidate neurons (filters) to be removed \cite{hu2016network}. The major issues of these methods include the difficulty of calculating the pruning ratio for each iteration and the large computation burden for the iterative sparse training and sensitivity analysis.

\textbf{Cost function Taylor-expansion-based methods. }Molchanov et al.\cite{Molchanov2017pruning} proposed a pruning criterion determination method by evaluating the first-order Taylor expansion of the loss function with respect to the features to be considered. The approach requires a time consuming sensitivity analysis at each pruning iteration to evaluate the consequence of deleting a specific parameter or filter. This method was later extended to identify candidate weak parameters based on the first and second order Taylor approximation of the loss function with respect to parameters \cite{Molchanov2019Taylor}. A significant issue of the Taylor expansion-based method is the ignorance of correlations among filters, resulting in a low efficient process by pruning the correlated filters one by one. 

\textbf{Feature reconstruction-based methods. }This type of method differs from the others by constraining the feature reconstruction fidelity during the pruning process. Filters can only be removed if their removal has the minimal feature reconstruction errors compared to the cases of removing other filters in the same layer \cite{Zhang2015Approximations, luo2017thinet, He2017Accelerating,yu2018nisp}. This approach assumes a low-rank of the feature map space, which may not be accurate and the resultant errors will be propagated into and enlarged in successive layers. Yu et al. proposed a method to partially mitigate this error accumulation issue\cite{yu2018nisp}. For deep neural network, this feature space decomposition based filter removing and network structure adjustment process will be computationally exhaustive. This problem can be partially addressed using a generative adversarial learning-based pruning but at the expense of extra optimization process and the hard filter pruning used therein may be ineffective and lacks flexibility\cite{Lin2019Generative}. 

\textbf{The information-theory-based methods. }This type of methods are designed to achieve the maximum pruning ratio while keeping the parameter diversity or output diversity of the network. Similar data features captured by different filters increase the network redundancy and can be pruned away as they do not provide extra information for the data distribution learned by the network \cite{Luo2017Entropy}. Based on the fact of that many filters have similar contributions even though their norms are small, He et al. \cite{He2019Geometric} proposed the geometric median as the pruning criterion instead of the contribution strength. Ding et al. extended this median based pruning approach into a multiple clusters-based approach \cite{Ding2019CSGD}. A slightly different approach was proposed by Lin et al. \cite{Lin2019HRank} based on the rank of feature maps. The diversity or information based methods heavily depend on the data as either the feature maps or the neuron output are required, which inevitably needs long computation time to calculate the diversity of features or output.

\section{Theory and methods}

\subsection{Preliminaries}
Our focus in this paper was the convolutional neural network (CNN). A CNN can be recursively described through:
\begin{equation}
\begin{small}
\label{eq1}
    {x}_{l+1} = {f}({\delta}_{l+1}({x}_{l},\{{W}_{l+1}\}))\quad\quad    {l}=0,1,..., {L-1},
\end{small}
\end{equation}
where f is the nonlinear operator, which is the Rectified Linear Unit (ReLU) in this paper; ${\delta}_{l+1}$ is the convolution operator; ${x}_{l}$ is the input tensor of the $l_{th}$ convolutional layer, with tensor shape $\left \langle {c}_{l}, {h}_{l}, {w}_{l} \right \rangle$ with ${h}_{l}$ and ${w}_{l}$ indicating the spatial dimension and ${c}_{l}$ the channel dimension; ${W}_{l+1}$ is a linear projection, with tensor shape $\left \langle {c}_{l+1}, {c}_{l}, {k}, {k} \right \rangle$, where ${c}_{l+1}$ and ${c}_{l}$ are channel dimension and ${k}$ is the convolution kernel size. To simplify the description of permutation symmetry analysis, a ${1}\times{1}$ kernel size was assumed and batch-normalization was skipped in the following text.

Weights of each convolutional layer are often initialized to be independently identically distribution. According to the chain rule, the derivative of the loss function with respect to a single weight ${W}_{l,i,j}$ between two adjacent convolutional layers ${l-1}$ and ${l}$ is given by:
\begin{small}
\begin{equation}
\label{eq2}
\begin{split}
    \frac{\partial\,{E}}{\partial\,{W}_{l,i,j}}
    &=\frac{\partial\,{E}}{\partial\,{x}_{l,j}}\frac{\partial\,{x}_{l,j}}{\partial\,{f}_{l,j}}\frac{\partial\,{f}_{l,j}}{\partial\,{\delta}_{l,j}}\frac{\partial\,{f}_{l,j}}{\partial\,{W}_{l,i,j}}\\
    &=\frac{\partial\,{E}}{\partial\,{x}_{l,j}}\frac{\partial\,{x}_{l,j}}{\partial\,{f}_{l,j}}\frac{\partial\,{f}_{l,j}}{\partial\,{\delta}_{l,j}}\sum_{m=1}^{{h}_{l-1}} \sum_{n=1}^{{w}_{l-1}}{x}_{l-1,i,m,n}\,,
\end{split}
\end{equation}
\end{small}
where E($\cdot$) is the error function, $i$ is the index of feature channels at layer ${l-1}$,\, $j$ is the index of feature channels at layer ${l}$,\, ${h}_{l-1}$ and ${w}_{l-1}$ are the height and width of feature maps at layer ${l-1}$.

Below is an introduction to the three types of singularities in CNN \cite{orhan2018skip}.

\textbf{Overlap Singularities}: Overlap singularity is induced by the permutation symmetry of hidden nodes. Let's consider a different connection ${W}_{l,i',j}$ between the same feature channel ${j}$ at layer ${l}$ and a different feature channel ${i'}$ at layer ${l-1}$. If ${x}_{l-1,i}={x}_{l-1,i'}$, then $\sum_{m=1}^{{h}_{l-1}} \sum_{n=1}^{{w}_{l-1}} {x}_{l-1,i,m,n}=\sum_{m=1}^{{h}_{l-1}} \sum_{n=1}^{{w}_{l-1}} {x}_{l-1,i',m,n}$ for all possible inputs, all the remaining terms in Equation (\ref{eq2}) are independent of the index $i$. Thus, the derivative of the cost function with respect to ${W}_{l,i,j}$ becomes identical to its derivative with respect to ${W}_{l,i',j}$ : $\frac{\partial\,{E}}{\partial\,{W}_{l,i,j}}=\frac{\partial\,{E}}{\partial\,{W}_{l,i',j}}$. In this condition of ${x}_{l-1,i}={x}_{l-1,i'}$, if the sum of ${W}_{l,i,j}$ and ${W}_{l,i',j}$ is satisfied by a constant value c : ${W}_{l,i,j}+{W}_{l,i',j}=c$, then, ${W}_{l,i,j}{x}_{l-1,i}+{W}_{l,i',j}{x}_{l-1,i'}=c{x}_{l-1,i}$ for any ${W}_{l,i,j}$ and ${W}_{l,i',j}$. During training, these parameters ${W}_{l,i,j}$ and ${W}_{l,i',j}$ will lose identifiability, the output of the model will be changed by the sum of ${W}_{l,i,j}$ and ${W}_{l,i',j}$ instead of the value of each ${W}_{l,i,j}$ and ${W}_{l,i',j}$.

\textbf{Linear singularities:} Similarly to overlap singularities, the linearity-induced singularity is caused by the linear dependence of the inputs, making a set of units in the same layer linearly dependent. Let's consider a subset of different connections between the same feature channel ${j}$ at layer ${l}$ and a subset of different feature channels, such as ${i}$, ${i'}$, ${i''}$, at layer ${l-1}$. If ${x}_{l-1,i,m,n}$, ${x}_{l-1,i',m,n}$ and ${x}_{l-1,i'',m,n}$ are linearly dependent, the derivatives with respect to those parameters become linearly dependent, thereby making the Hessian singular. During training, only a linear combination of them is identiﬁable.

\textbf{Elimination Singularities:} The elimination singularity is caused by the zero incoming weight or zero input, which actually indicates that the corresponding units are completely ineffective and should be removed. If the parameters $W_{l, i, :}$ in the $l$ convolutional layer and connected to the i-th channel  of $l-1$ layer is 0, the filter parameters $W_{l-1, :, i}$ corresponding to the i-th channel in the previous convolutional layer $l-1$ are no longer discriminative, and the final output value of the model is regardless of the value of parameter $W_{l-1, :, i}$. Similarly, if $W_{l-1, :, i}=0$, then the parameters connected with the i-th channel of the next convolutional layer $l$ will loss identity. No matter what value $W_{l, i, :}$ is taken, the final output value of the model will not be affected.

\subsection{Hessian matrix analysis}

As Dong et al.\cite{LayerWiseOBS2017NIPS} denoted, the layer-wise pruning error can be described as an error function $PE(\cdot)$:
\begin{equation}
\begin{small}
\label{eq3}
    PE(\widetilde{Z_{l}})=\frac{1}{n}\left \| \widetilde{Z_{l}}-Z_{l} \right \|_{F}^{2} \; \; l=0,1......,L-1,
\end{small}
\end{equation}
where $Z_l$ denotes the outcome of convolution operation before performing the activation function at well-trained convolution layer $l$, $\widetilde{Z_{l}}$ denotes the output feature map of convolution layer l after pruning at convolution layer $l$, $\left \| \cdot  \right \|_{F}$ denotes Frobenius norm.

The layer-wise pruning error function $PE(\widetilde{Z_{l}})$ can be approximated by Taylor expansion as follow:
\begin{small}
\begin{equation}
\begin{split}
\label{eq4}
    PE(\widetilde{Z_{l}})= PE(Z_{l})\: 
    &+ \: \frac{\partial PE(Z_{l})}{\partial W_{l}}\Delta W\: \\
    &+ \: \frac{1}{2}\Delta W_{l}^{T} H_{l} \Delta W\: \\
    &+ \: O\left ( \left \| W_{l} \right \|^{3} \right )\; \; l=0,1,......,L-1,
\end{split}
\end{equation}
\end{small}
where $\Delta W_{l}$ denotes the difference of variable $W_{l}$ before and after pruning, $H_{l}=\frac{\partial^2 PE(Z_{l})}{\partial W_{l}^2}$, $O\left ( \left \| W_{l} \right \|^{3} \right )$ denotes the third and all higher order items of Taylor expansion formula. Obviously, $PE(Z_{l})=0$.  For a well-trained neural network, $\frac{\partial PE(Z_{l})}{\partial W_{l}}\Delta W$ is equal to zero, and $O\left ( \left \| W_{l} \right \|^{3} \right )$ can be ignored.

So far, the pruning error of each convolution layer after pruning can be formally expressed as the following optimization problem:
\begin{small}
\begin{equation}
\label{eq5}
    min PE(\widetilde{Z_{l}})= min \frac{1}{2}\Delta W_{l}^{T} H_{l} \Delta W
    \; \; l=0,1,......,L-1.
\end{equation}
\end{small}

And then, let's conduct a Singular Value Decomposition (SVD) for $H_{l}$:
\begin{small}
\begin{equation}
\begin{split}
\label{eq6}
    H_{l}&= \sum_{i=1}^{r}\sigma _{i}u_{i}v_{i}^T \\
         &= \sum_{i=1}^{r^{'}}\sigma _{i}u_{i}v_{i}^T + \sum_{i=r^{'}+1}^{r}\sigma _{i}u_{i}v_{i}^T
    \; \; l=0,1,......,L-1,
\end{split}
\end{equation}
\end{small}
where $r$ denotes the rank of Hessian matrix $H_{l}$ and $r^{'}<r$, $u_{i}$ denotes the top-i singular values, $\sigma _{i}$ denotes the corresponding left singular vector, $v_{i}$ denotes the corresponding right singular vector. 

Differ from OBS\cite{OBS1992NIPS} and L-OBS\cite{LayerWiseOBS2017NIPS}, our method does not calculate the inverse Hessian matrix for each layer. We merely focus on analysing the singular value components of Hessian matrix. As mentioned above, because of three singularities existing in training process, the rank of Hessian matrix $H_{l}$ is not full. Also, the equation \ref{eq6} shows us that Hessian matrix $H_{l}$ with rank $r$ can be decomposed into a lower-rank matrix with rank $r_{'}$ ($\sum_{i=1}^{r^{'}}\sigma _{i}^Tu_{i}v_{i}$) and the other matrix with some additional information ($\sum_{i=r^{'}+1}^{r}\sigma _{i}^Tu_{i}v_{i}$). Combined with equation \ref{eq5} and \ref{eq6}, it can be seen that the effect of deleting some small singular value components on pruning error is relatively small. The number of small singular value components corresponds to the number of filters to be pruned. So this paper wants to set the proportion of singular value components to guide the calculation of the number of filters to be pruned.

\subsection{PCA on the Gradient Matrix}
Since Hessian matrix is computationally expensive, this section mainly transforms the above analysis of Hessian matrix into using Principle Component Analysis(PCA) on gradient matrix of neural network parameters through formula derivation.

Below we proved that the Hessian matrix of loss function is directly related to the expectation of the covariance of the gradient. Let us consider the loss function as the negative logarithm of the likelihood. Let $X$ be a set of samples and $p(x; \theta)$ be the distribution over $X$, which can be implicitly described by a parameterized neural network associated with $\theta$. The Fisher information of the set of probability distributions $P = \left \{p(x; \theta): \theta \in \Theta \right \}$ can be described by a matrix whose value at the $i$-th row and $j$-th column is:
\begin{equation}
\begin{small}
\label{eq7}
    I_{i,j}(\theta )=E_{X}[\frac{\partial log p(x;\theta )}{\partial \theta_{i}}\frac{\partial log p(x;\theta )}{\partial \theta_{j}}].
\end{small}
\end{equation}

It is then trivial to prove that the Fisher information of the set of probability distributions P approaches a scaled version of the Hessian of log likelihood: 
\begin{equation}
\begin{small}
\label{eq8}
    I_{i,j}(\theta )= -E_{X}[\frac{\partial^2 log p(x;\theta )}{\partial \theta_{i} \partial \theta_{j}}].
\end{small}
\end{equation}

Let $D_{i}log p(x; \theta)$ denote the first-order partial derivatives $\frac{\partial }{\partial \theta_{i}}$, $D_{i, j}$ denote the second-order partial derivative $\frac{\partial^2 }{\partial \theta_{i}\partial \theta_{j}}$. Noting that the first derivatives of log likelihood is
\begin{equation}
\begin{small}
\label{eq9}
    D_{i}log p(x; \theta) = \frac{D_{i}p(x;\theta)}{p(x;\theta)}.
\end{small}
\end{equation}

And the second derivatives is 
\begin{small}
\begin{equation}
\label{eq10}
    D_{i, j}log p(x;\theta) = \frac{D_{i,j}p(x;\theta)}{p(x;\theta)}-
    \frac{D_{i}p(x;\theta)}{p(x;\theta)}\frac{D_{j}p(x;\theta)}{p(x;\theta)}.
\end{equation}
\end{small}

By taking the expectation of the second derivative and using the trick that the second derivatives and integrals can be switched, we can obtain:
\begin{small}
\begin{equation}
\begin{split}
\label{eq11}
E_{X}(D_{i, j}log p(x, \theta))
 &=  -E_{X}\left \{\frac{D_{i}p(x,\theta)}{p(x,\theta)}\frac{D_{j}p(x,\theta)}{p(x,\theta)}\right \} \\ 
 &=  -E_{X}\left \{D_{i}log p(x,\theta))(D_{j}log p(x,\theta)\right \},
\end{split}
\end{equation}
\end{small}
where,
\begin{small}
\begin{align*}
 E_{X}(\frac{D_{i, j}p(x,\theta)}{p(x,\theta)})
 &= \int \frac{D_{i,j}p(x;\theta)}{p(x;\theta)} p(x, \theta)dx\\ 
 &= D_{i, j}\int p(x, \theta)dx\\ 
 &= 0.
\end{align*}
\end{small}
This concludes the proof of the connection between the covariance of gradients and the Hessian of the log likelihood. The following PCA-based network redundancy estimation method was based on this Hessian matrix vs gradient matrix relationship. 

Another way to establish a connection between Hessian matrix and gradient matrix is to deduce from equation \ref{eq3}. For each convolution layer l, the first derivative of the pruning error function with respect to $W_{l}$ is $\frac{\partial Z_{l}}{\partial W_{l}}\left ( \widetilde{Z_{l}}-Z_{l} \right )$, and the Hessian matrix is defined as: $\frac{\partial Z_{l} }{\partial W_{l}}\left ( \frac{\partial Z_{l}}{\partial W_{l}} \right )^T - \frac{\partial^2 Z_{l}}{\partial W^2}\left ( \widetilde{Z_{l}} - Z_{l} \right )^T$. As OBS pointed out\cite{OBS1992NIPS},  for most cases $\widetilde{Z_{l}}$ is close to $Z_{l}$. So Hessian matrix can be calculated by $\frac{\partial Z_{l} }{\partial W_{l}}\left ( \frac{\partial Z_{l}}{\partial W_{l}} \right )^T$.

After interpreting the connection between Hessian matrix and gradient matrix of parameters, next step is to explain why we introduce a preprocessing method to avoid some stochastic errors. Stochastic error means the gradient calculated by a batch of sample is different from the gradient calculated by all samples. Inspired by several previous studies \cite{Molchanov2017pruning,Molchanov2019Taylor,Ding2019GSM}, we used the first-order Taylor expansion to suppress the gradients of some parameters that may be affected by the stochastic errors during the random mini-batches based network training. After this preprocessing, the stochastic errors to these parameters will have little effect on the Hessian degeneration. At each training iteration with a mini-batch of examples ${X}^{i}$ and labels ${Y}^{i}$, let ${T}({X}^{i},{Y}^{i},{w})$ be the impact of a specific parameter ${w}$ to the loss function, which will be 0 when the change of $w$ does not change the loss function.
\begin{small}
\begin{equation}
\begin{split}
\label{eq12}
    {T}({X}^{i},{Y}^{i},{w})&=\left | E(X^{i},Y^{i},W|w=0) - E(X^{i},Y^{i},W)\right |.
\end{split}
\end{equation}
\end{small}
While the value of $E(X^i,Y^i, W|w=0)$ is unknown, we can use the first order Taylor expansion formula to estimate $E(X^i,Y^i,W|w=0)$ and get the following equation:
\begin{small}
\begin{equation}
\begin{split}
\label{eq13}
    E(X^i,Y^i,W|w=0) = & E(X^i,Y^i,W) \\
                     & + \frac{\partial E({X}^{i},Y^i,W)}{\partial w}(0-w)+o(w^2).
\end{split}
\end{equation}
\end{small}
Then, we can get:
\begin{small}
\begin{equation}
\begin{split}
\label{eq14}
    {T}({X}^{i},{Y}^{i},{w})&=\left | \frac{\partial E({X}^{i},Y^i,W)}{\partial w} (0-w)\right |.
\end{split}
\end{equation}
\end{small}
From this equation, we can see that ${T}({X}^{i},{Y}^{i},{w})=0$ mean the change of $w$ has no impact to loss function as $w$ and $\frac{\partial E({X}^{i},Y^i,W)}{\partial w}$ are close to zero. In this case, the gradient of $w$ is set to be zero.

\begin{figure*}[tbp]
\begin{center}
    \includegraphics[width=0.8\linewidth]{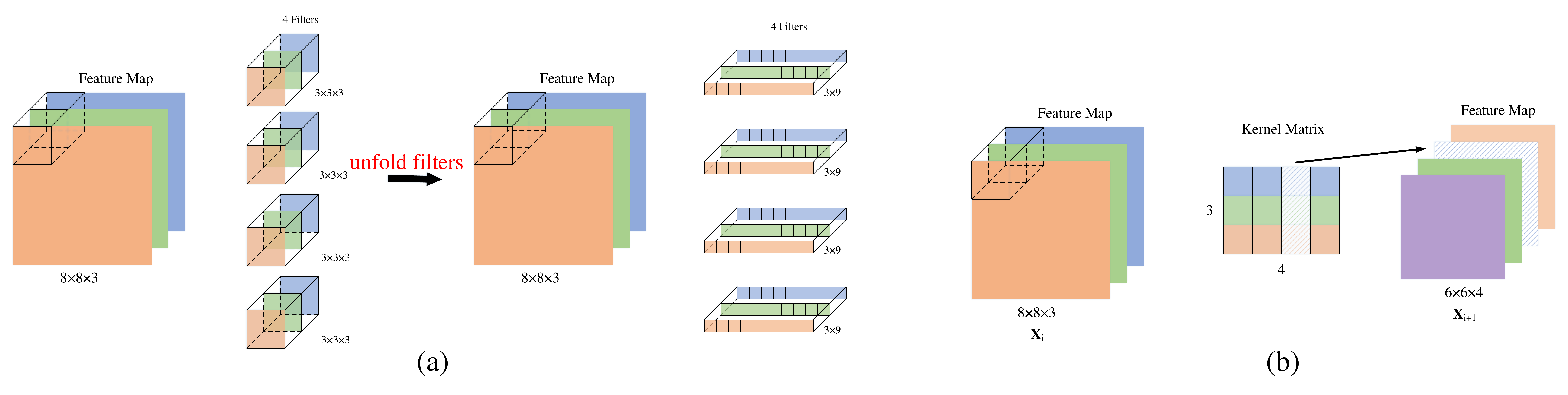}
\end{center}
    \caption{(a) illustrates the feature learning in an standard convolutional neuronal network, each filter with 3x3 kernel size is unfolded into 1x9 kernel size, and (b) illustrates the corresponding feature maps and the kernel matrix. Permutation symmetry does not exist within the channel because of the limited receptive ﬁeld size and weight sharing, but exists across feature channels.}
\label{Channel_Dimension}
\end{figure*}

After zeroing out some of the weight gradients, we applied PCA\cite{Hotellings1933PCA} decomposition on the weight gradient matrix: ${[\frac{\partial\,{E}}{\partial\,{W}_{l}}]}_{cin \times cout}$ for each layer separately, where $l$ is the index of the convolutional layer, $cin$ is the number of input channels, $cout$ is the number of the output channels. The weight gradient matrix is the gradient matrix of the kernel matrix(illustrated in Fig.\ref{Channel_Dimension}). Generally, there is a doubt about the manually set filter pruning ratio. If the filter pruning ratio is set to be 50\%, we can not confirm how much redundant information the 50\% filters removed from model contain. PCA provides a variance contribution rate to denote the amount of variation captured by PCA directions. Therefore, in this paper, a global variance contribution rate is used to calculate the `filter pruning ratio' of the model, which can be used to analyze each convolutional layer. For example, a 90\% global variance contribution ratio means that the retained filters, after reducing the number of original filters, contain 90\% of the original information of a convolutional layer. Obviously, the smaller the global variance contribution rate is, the larger the filter pruning ratio is. Given the global variance contribution rate, we can determine the number of non-dominant eigenvectors for each layer. Then, the ratio of the sum of all layers' non-dominant eigenvector numbers relative to the total number of all possible eigenvectors is defined as the 'filter pruning ratio'.

We call this PCA-based network redundancy estimation method `Effective Filter Analysis'(\textbf{EFA}).

\subsection{Layer-wise Pruning Error and the Layer Overpruning Problem}

As Dong et al. discussed in their paper\cite{Dong2017Surgeon}, the accumulated error of ultimate network output $\widetilde{\varepsilon }_{L-1}=\frac{1}{\sqrt{n}}\left \| \widetilde{Y}_{L-1}-Y_{L-1} \right \|_{F}$ obeys:
\begin{small}
\begin{equation}
\label{eq15}
    \widetilde{\varepsilon }_{L-1}\leq =\sum_{k=1}^{L-2}\left ( \prod_{l=k+1}^{L-1} \left \| \widehat{W}_{l} \right \|_{F} \sqrt{\Delta PE_{k}} \right )+\sqrt{\Delta PE_{L-1}},
\end{equation}
\end{small}
where $\widetilde{Y}_{L-1}$ is the accumulated pruned output of the layer $L-1$ after performing activation function $f (\cdot )$. $\widehat{W}_{l}$ denotes the new parameter vector of layer $l$ after pruning. $\Delta PE$ denotes a perturbation of pruning error before and after pruning. $1\sim {L-1}$ denotes the direction from input to output. $F(\cdot)$ denotes frobenius norm.

Equation \ref{eq15} shows that layer-wise pruning errors will be scaled by continued multiplication of parameters’ Frobenius Norm over the following layers. As mentioned above, pruning the filters across all layers simultaneously has a potential issue which is some layers get over-pruned (having more filters removed than others). If some internal layers get over-pruned and only remain one or two filters, pruning errors in those layers will be scaled by continued multiplication of parameters’ Frobenius Norm over the following layers. The final result is likely to be the abrupt network performance decline.

\subsection{The Hierarchical Pruning Algorithm}

\begin{figure}[tbp]
 \begin{center}
     \subfigure[]{
         \includegraphics[width=0.9\linewidth]{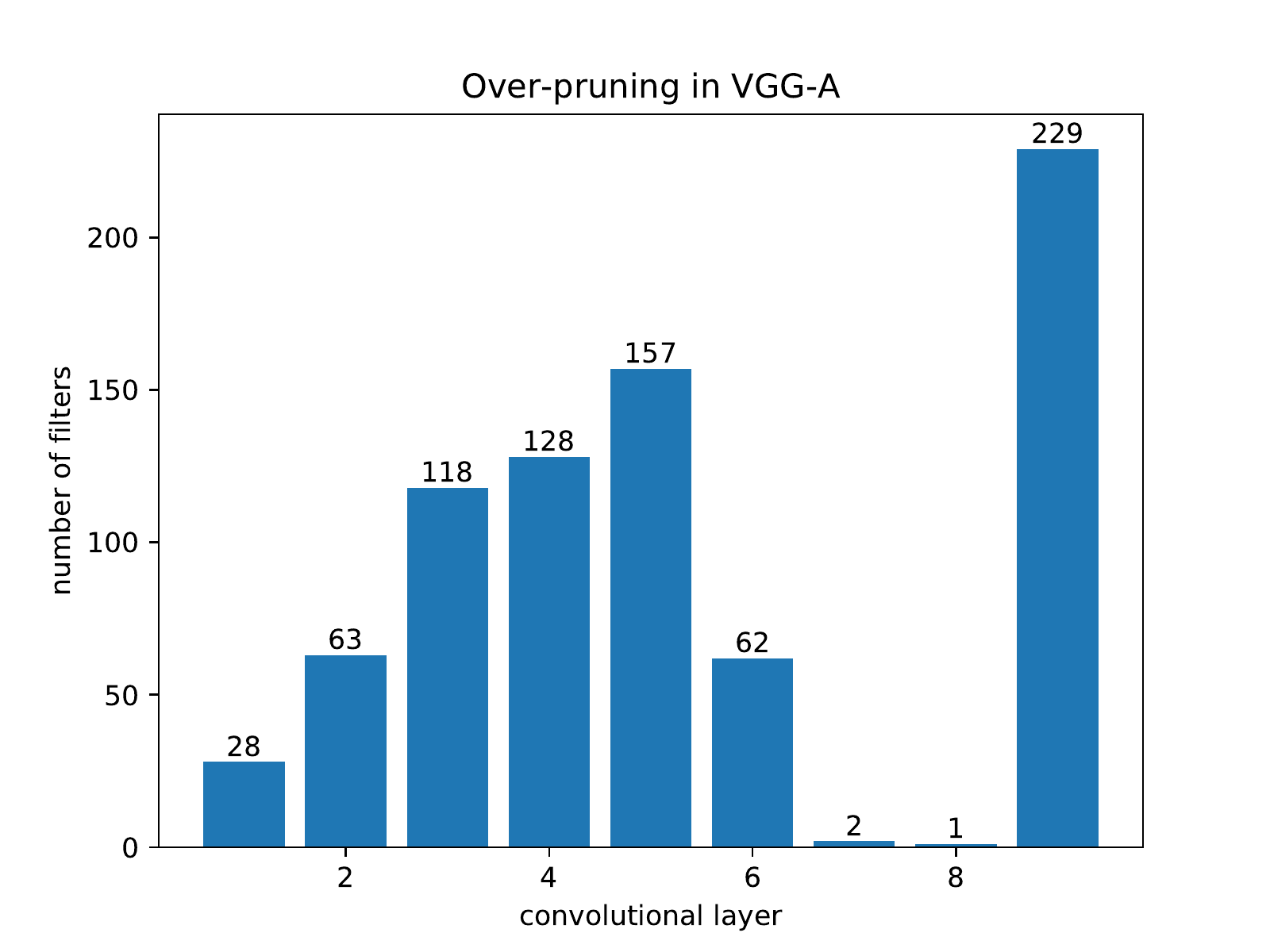}
     }
     \subfigure[]{
         \includegraphics[width=0.9\linewidth]{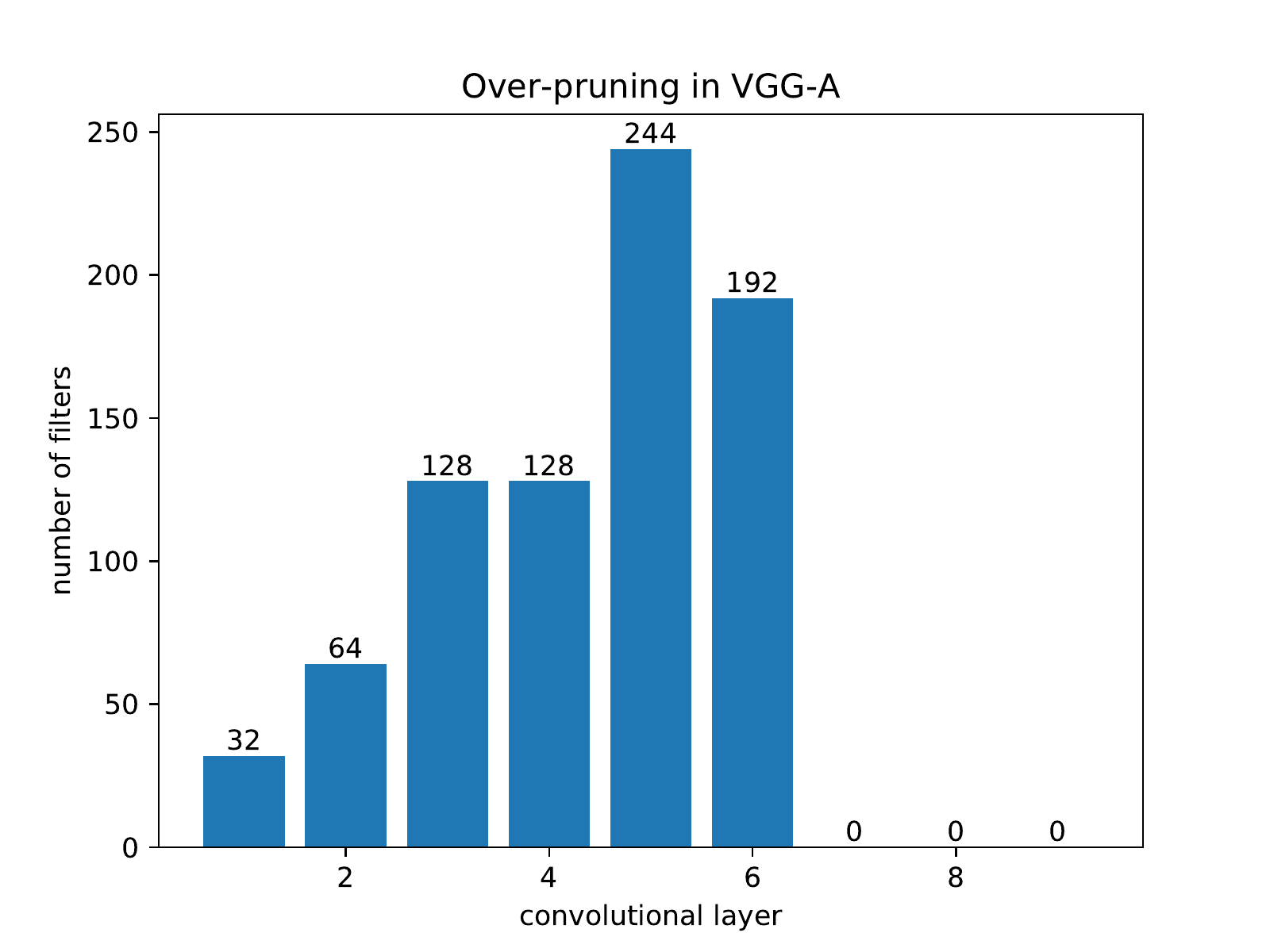}
     }
 \end{center}
 \caption{Filter distribution patterns of VGG-A after filter pruning using the traditional filter-oriented pruning method and our new hierarchical method.}
 \label{OverPruning}
 \end{figure}

After determining the `filter pruning ratio' via the EFA method, we used the following hierarchical algorithm to prune the network. Our pruning algorithm is still an iterative process. It is worth noting that instead of ﬁne-tuning the network every time after pruning a convolution layer, we prune the network in a one-shot way. To avoid over-pruning problem\ref{OverPruning}, at each iteration, the pruning step contains a layer-wise and a filter-wise pruning. Each of the substeps depends on a separate layer or filter selection criterion. More pruning details are showed on algorithm \ref{algorithm 1}. 

\noindent\textbf{Layer selection criterion.} We used cross entropy of two adjacent layers to find the candidate layer to be pruned at each step. Cross entropy measures the similarity of the distribution of the weights of two layers. Lower cross entropy means that the two adjacent layers are statistically similar to each other up to certain linear scaling. The probability distribution of the weights of each layer was estimated based on the histogram of the weights. For each layer, the weights were normalized to a norm of 1. The histogram was calculated by grouping the weights into m different bins, and calculating the proportion of the number of weights in each bin in relative to the total number of weights. The cross entropy of adjacent layers was then calculated by:
\begin{small}
\begin{equation} \label{eq:crossen}
    CrossEntropy = -\sum_{j=1}^{m}\,p_{j}^{L}\,log\, p_{j}^{L-1},
\end{equation}
\end{small}
where, $p_{j}^{L}$ is the probability of j-th bin in layer $L$. Layers with more similar probability distributions of weights will have smaller cross entropy.

\noindent\textbf{Filters selection criterion:} For each filter, we divide the weight value into m different bins (m=1000 in this paper), and calculate the probability of each bin. The information entropy of each filter can be calculated as follows:
\begin{small}
\begin{equation} \label{eq:filterentropy}
    {H}_{i}^{L} = -\sum_{j=1}^{m}\,p_{j}\,log\,p_{j}\quad\quad {i} = 1,\cdots ,{C}^{L},
\end{equation}
\end{small}
where, $p_{j}$ is the probability of the j-th bin, ${H}_{i}^{L}$ is the entropy of the i-th filter in layer $L$, ${C}^{L}$ is the number of filters in layer $L$. In general, the smaller the information entropy is, the more single the parameter values are.

The pruning threshold on information entropy could be calculated by the `filter pruning ratio'. For example, 40\% filters are pruned, the information entropy of all filters of the whole model are arranged in descending order, and the information entropy threshold is in the 40-th percentile counting backwards of the sorted information entropy of all filters.
\begin{algorithm}[ht]
    \caption{The hierarchical layer and filter pruning algorithm}
    \label{algorithm 1}
    \KwIn{A trained model}
    \KwOut{A pruned model}
    Estimate the filter pruning ratio by our EFA method, then calculate the total number of remaining filters ,$N$ \;
    Calculate the information entropy of each filters at different layer: $Info$ \;
    Calculate the cross entropy of two successive layers of model at different layer: $CE$ \;
    $flag\leftarrow True$ \;
    \While{flag}{
        $SortedInfo\leftarrow sort(Info, descending)$ \;
        Calculate the  global threshold across all layers, $thre\leftarrow SortedInfo[N]$ \;
        Select N filters by information-entropy filter selection criterion\;
        \eIf{the number of remaining filters in a layer are less than a certain number}
        {
            Zero out the layer selected by cross-entropy layer selection criterion\;
            Zero out the information entropy of the corresponding filters from $Info$\;
        }{
            $flag\leftarrow False$ \;
        }
    }
\end{algorithm}

\begin{table*}[tbp]
\centering
\fontsize{8}{10}\selectfont
\begin{tabular}{|l|l|l|l|l|}
\hline
Model                  & Accuracy & Filters(PR)     & Parameters(PR)    & FLOPs(PR)         \\ \hline
VGG-A(baseline)        & 93.69\%  & 4224(-)         & 14.7M(-)          & 314M(-)           \\ \hline
VGG-A(pruned\_0.99)    & 93.57\%  & 1287(69.53\%)   & 1.83M(87.55\%)    & 183.22M(41.65\%)  \\ \hline
VGG-A(pruned\_0.95)    & 93.49\%  & 788(81.34\%)    & 0.78M(94.69\%)    & 125.84M(59.92\%)  \\ \hline
VGG-A(pruned\_0.9)     & 91.36\%  & 568(86.55\%)    & 0.47M(96.80\%)    & 106.54M(66.07\%)  \\ \hline
ResNet56(baseline)     & 93.73\%  & 1008(-)         & 0.85M(-)          & 126.80M(-)        \\ \hline
ResNet56(pruned\_0.99) & 93.46\%  & 629(36\%)       & 0.44M(48.24\%)    & 96.61M(23.81\%)   \\ \hline
ResNet56(pruned\_0.95) & 92.34\%  & 453(54\%)       & 0.24M(71.76\%)    & 55.77M(56.02\%)   \\ \hline
ResNet56(pruned\_0.9)  & 91.92\%  & 355(64\%)       & 0.17(80.00\%)     & 49.34M(61.09\%)   \\ \hline
\end{tabular}
\caption{Network pruning evaluation results based on CIFAR-10. Accuracy means the top-1 accuracy. PR means pruned ratio. Baseline means the original model without pruning. Pruned\_0.99, Pruned\_0.95 and Pruned\_0.9 mean that the number of filters of pruned model was calculated by the global variance contribution ratio 0.99, 0.95 and 0.9, respectively. M means million(1e6).}
\label{table 1}
\end{table*}

\begin{table*}[tbp]
\centering
\fontsize{8}{10}\selectfont
\begin{tabular}{|l|l|l|l|}
\hline
Method                                  & Top-1   & Parameters(PR$\uparrow$)    & FLOPs(PR)         \\ \hline
ResNet56                                & 93.73\% & 0.85M(-)                    & 126.80M(-)        \\ \hline
\textbf{ResNet56 (pruned\_0.99, ours)}  & 93.46\% & 0.44M(48.24\%)              & 96.61M(23.81\%)   \\ \hline
HRank\cite{Lin2019HRank}                & 93.17\% & 0.49M(42.4\%)               & 62.72M(50.53\%)    \\ \hline
\textbf{ResNet56 (pruned\_0.95, ours)}  & 92.34\% & 0.24M(71.76\%)              & 55.77M(56.02\%)   \\ \hline
\textbf{ResNet56 (pruned\_0.9, ours)}   & 91.92\% & 0.17M(80.00\%)              & 49.34M(61.09\%)   \\ \hline
HS\cite{He2017Accelerating}             & 90.8\%  & (-)                         & 62.00M(51.10\%)    \\ \hline
GAL-0.8\cite{Lin2019Generative}         & 90.36\% & 0.29M(65.88\%)              & 49.99M(60.58\%)    \\ \hline
HRank\cite{Lin2019HRank}                & 90.72\% & 0.27M(68.24\%)              & 32.52M(74.35\%)    \\ \hline
\end{tabular}
\caption{Pruning performance comparisons between different methods for the CIFAR-10 dataset.}
\label{table 2}
\end{table*}

\begin{table*}[tbp]
\centering
\fontsize{8}{10}\selectfont
\begin{tabular}{|l|l|l|l|l|l|}
\hline
Model                       & Top-1     & Top-5       & Filters(PR)   & Parameters(PR)  & FLOPs(PR)       \\ \hline
VGG16-bn(baseline)          & 73.36\%   & 91.52\%     & 4224(-)       & 138.37M(-)      & 15.53B(-)       \\ \hline
VGG16-bn(pruned\_0.99)      & 71.62\%   & 90.57\%     & 2535(39.99\%) & 37.66M(72.78\%) & 10.50B(32.39\%) \\ \hline
VGG19-bn(baseline)          & 73.99\%   & 91.69\%     & 5504(-)       & 143.68M(-)      & 19.69B(-)       \\ \hline
VGG19-bn(pruned\_0.99)      & 71.78\%   & 90.49\%     & 2590(52.94\%) & 29.09M(79.75\%) & 8.48B(56.93\%)  \\ \hline
ResNet50(baseline)          & 76.13\%   & 92.86\%     & 3776(-)       & 25.55M(-)       & 4.111B(-)       \\ \hline
ResNet50(pruned\_0.99)      & 73.21\%   & 91.11\%     & 1737(53.99\%) & 12.07M(52.76\%) & 1.721B(58.14\%) \\ \hline
\end{tabular}
\caption{Network pruning evaluation results based on ILSVRC-2012. Accuracy is measured by the top-1 and top-5 accuracy. PR means pruned ratio. Baseline means the original model without pruning. Pruned\_0.99 means that the number of filters of pruned model was calculated by the global variance contribution ratio 0.99. M/B means million/billion (1e6/1e9).}
\label{table 3}
\end{table*}

\begin{table*}[tbp]
\centering
\fontsize{8}{10}\selectfont
\begin{tabular}{|l|l|l|l|l|l|}
\hline
Method                                      & Top-1   & Top-5   & Parameters(PR)   & FLOPs(PR)          \\ \hline
ResNet50                                    & 76.13\% & 92.86\% & 25.55M(-)        & 4.111B(-)          \\ \hline
Taylor-FO-BN-72\%\cite{Molchanov2019Taylor} & 74.50\% & -       & 14.2M (44.42\%)  & 2.25B(45.26\%)     \\ \hline
\textbf{ResNet50( Pruned\_0.99, ours)}      & 73.01\% & 91.11\% & 12.07M (52.76\%) & 1.721B(58.14\%)    \\ \hline
NISP-50-B\cite{yu2018nisp}                  & 72.07\% & -       & 14.3M (44.03\%)  & 2.29B(44.29\%)     \\ \hline
HRank\cite{Lin2019HRank}                    & 71.98\% & 91.01\% & 13.77M (46.11\%) & 1.55B(62.30\%)     \\ \hline
Taylor-FO-BN-56\%\cite{Molchanov2019Taylor} & 71.68\% & -       & 7.9M (69.08\%)   & 1.34B(67.40\%)     \\ \hline
ThiNet\cite{luo2017thinet}                  & 71.01\% & 90.02\% & 12.38M (51.55\%) & 1.71B(58.40\%)     \\ \hline
GAL-1\cite{Lin2019Generative}               & 69.88\% & 89.75\% & 14.67M (42.58\%) & 1.58B(61.57\%)     \\ \hline
\end{tabular}
\caption{Pruning performance comparison results of different pruning methods on ILSVRC-2012.}
\label{table 4}
\end{table*}

\section{Experiments}
\subsection{Datasets and Network Models}
\noindent\textbf{Datasets.} Our pruning framework was evaluated on two benchmark datasets:  CIFAR-10\cite{Hinton2009Cifar} and ILSVRC-2012\cite{Fei2015Imagenet}. CIFAR-10 contains 60,000 $32 \times 32$ natural images categorized into 10 classes. 50,000 were used for model training; 10,000 were used for testing. ILSVRC-2012 has 1.33 million images classified into 1000 classes: 1.28 million was used for training, 50k for validations. 

\noindent\textbf{Deep networks. }The hierarchical pruning algorithm was tested with VGGNets\cite{VggNet2015ICLR} and ResNets\cite{ResNet2016CVPR} with residual blocks. The following networks that are often used in network compression experiments were assessed: VGG16-bn, VGG19-bn, ResNet56, ResNet50. VGG16-bn and ResNet56 were trained and tested with CIFAR-10. VGG16-bn, VGG19-bn, and ResNet50 were trained and tested using data in ILSVC-2012.

VGG16-bn is originally designed for ImageNet classification. In this study, the original VGG16-bn network architecture was changed to fit the need of CIFAR-10 dataset. The new structure, which we called VGG-A consists of 13 convolutional layers and a fully connected layer. Each convolutional layer has a batch normalization (bn) layer \cite{BN2015ICML} inserted before the activation function. 

ResNet56 has fewer parameters than VGG-A and is more challenging to prune. ResNet56 has three stages of residual blocks for outputting feature maps with sizes of $32\times32$, $16\times16$ and $8\times8$. Each stage contains the same number of residual blocks. The residual block is comprised of two convolutional layers with a kernel size of $3\times3$ and a shortcut layer. The shortcut layer provides an identity mapping with an additional zero padding for the increased dimensions and does not need pruning. In other words, only the ﬁrst layer of the residual block of each stage was pruned. 

VGG16-bn for ILSVRC-2012 consists of 13 convolutional layers and 3 fully connected layers. VGG19-bn for ILSVRC-2012 consists of 16 convolutional layers and 3 fully connected layers. Each convolutional layer has a batch normalization layer next to it. Each fully connected layer is connected to a dropout layer\cite{Srivastava2014dropout}. To prune the neurons in fully-connected layers, we treated them as convolutional channels with $1\times1$ spatial size. Dropout has a tunable hyperparameter p (the probability of retaining a unit in the network). Before pruning, we set p=0.5. After pruning, we set p=0.8. For ILSVRC-2012, ResNet50 is used to prove that our method can work well on a multi branch structure. ResNet50 for ILSVRC-2012 contains a convolutional layer with $3\times3$ filters, a batch normalization layer, four stages of residual blocks and a fully connected layer. The residual blocks of four stages output feature maps with sizes of $56\times56$, $28\times28$, $14\times14$ and $7\times7$, respectively. ResNet50 contains a 3-layer bottleneck block as residual block. The three layers are $1\times1$, $3\times3$ and $1\times1$ convolutions, where the $1\times1$ convolutional layers are responsible for reducing and then increasing (restoring) dimensions to match the identity mapping. Pruning was not applied to the first convolutional layer the and $1\times1$ ﬁlter in ResNet50 because they involved much less computation compared to the $n \times n$ ﬁlters. 

\subsection{Illustrating the Layer Overpruning Risk}
As noticed in previous studies, different layer contributes to network performance differently. Pruning a significant portion of filters of some layers may cause a sudden network performance drop. To illustrate this layer over-pruning risk and to demonstrate the efficacy of our new hierarchical pruning method, we compressed VGG-A using CIFAR-10 with a large pruning ratio. The prior-pruning rate was set to be 81\%. The "filter selection criteria" proposed in the "network slimming" method \cite{liu2017learning} was used as a comparison to our hierarchical layer-first-filter-second pruning process. The number of remaining filters in each layer was recorded to show the variety of filter pruning rate of each layer.  Network was re-trained after pruning. Fig. \ref{OverPruning} shows the filter pruning results based on the traditional filter-wise pruning method (the network slimming method \cite{liu2017learning}) and our hierarchical method. both methods showed very similar within-layer filter pruning rate pattern from the first to the 6-th layer. Using the traditional method, the 7-th and 8-th layer were substantially pruned with only one or two filters remained and the last layer was the least pruned layer. The compressed network yielded a test error of 90\%. By contrast our hierarchical method identified the last three layers to be the least contributing ones and completely removed them but only with a penality of a test error of 7.67\%.

\subsection{Network Implementation and Pruning} 
PyTorch\cite{paszke2017automatic} was used to implement all algorithms. Network training was based on the Stochastic Gradient Descent algorithm (SGD) with a 0.1 initial learning rate. For CIFAR-10, the batch size, weight decay and momentum were set to be 128, 0.0005 and 0.9, respectively. For ILSVRC-2012, 4 GPUs were used to train the models. The batch size, weight decay and momentum were set to be 256, 0.0001 and 0.9, respectively. The pruned models for CIFAR-10 were retrained from scratch for 160 epochs, with the learning rate divided by 10 in $\left\{80, 120 \right\}$ epochs. For ILSVC-2012, the pruned models were retrained from scratch for 100 epochs, with the learning rate divided by 10 every 30 epochs. For CIFAR-10, data augmentations through shifting/mirroring were used \cite{ResNet2016CVPR}. For ILSVRC-2012, data argumentations were performed using the options provided in PyTorch\cite{paszke2017automatic}. 

To prune the trained network, we first set the global variance contribution ratios to calculated the `filter pruning ratio' then started the pruning iteration. 

The following variance contribution ratios were used for VGG-A and ResNet56 respectively: $\left\{0.99, 0.95, 0.9 \right\}$. The corresponding filter pruning ratios calculated by our EFA method are listed on the third column of table \ref{table 1}. For VGG16-bn, VGG19-bn, ResNet50, the global variance contribution ratios were set to be 0.99. The corresponding filter pruning ratios calculated by our EFA method are listed on the third column of table \ref{table 3}. For VGGNets, the PCA decomposition was applied for the weight gradient matrix of each convolutional layer. For ResNet56/ResNet50, the PCA decomposition was applied for the weight gradient matrix of first/second convolutional layers of each residual block, respectively. Then, we removed filters from the corresponding convolutional layers. We counted the numbers of remaining filters of the corresponding layers of different models and list them in table \ref{table 1} and table \ref{table 3}.

At each pruning iteration, we first identified the layer with the least number of filters remaining (given a least number of filters: 5 filters) and having the lowest cross-entropy to the adjacent layers and removed the entire layer. If a convolutional layer was pruned, the weights and biases of the subsequent batch normalization layer were removed as well. If no layers were identified, we proceeded to remove the least contributing filters based on the within layer information entropy and quit the pruning iteration. When pruning iteration was finished, new models with fewer ﬁlters were created and retrained from scratch. 

\subsection{Network pruning method performance indices}
Model size was measured by the number of parameters. Float Points Operations (FLOPs) was used to measure the computational cost. To evaluate the task-speciﬁc capabilities, we recorded the top-1 classification accuracy of pruned models on CIFAR-10, top-1 and top-5 classification accuracy of pruned models on ILSVRC-2012. 
\subsection{Results}

\noindent\textbf{Method validation results using the CIFAR-10 data. } Table \ref{table 1} lists the method evaluation results using CIFAR-10. 
As expected, the pruning rates of filter, parameter, and FLOPs increased with the global variance contribution rate. The increase of pruning rate was accompanied by minor to moderate network prediction accuracy. In terms of negligible prediction accuracy loss and high pruning ratio, the best solution was to use a global variance contribution ratio of 0.95, which produced the `filter pruning ratio' of 81\% for VGG-A. 

Table \ref{table 2} shows the method comparison results. The ones labeled by ``ours" were those pruned using the new pruning method proposed in this paper. Three pruning methods were compared: the feature reconstruction-based methods (HS \cite{He2017Accelerating} and GAL-0.8\cite{Lin2019Generative}), and the information-theory-based method(HRank\cite{Lin2019HRank}). As compared to ResNet56(baseline), ResNet56(pruned\_0.99, ours) achieved a parameter compression ratio of 48.24\% and a loss of 0.27\% Top-1 accuracy; HRank pruned the total parameters by 42.35\% with a loss of 0.56\% Top-1 accuracy; ResNet56(pruned\_0.95, ours) achieved a parameter compression ratio of 71.76\% and a loss of 1.39\% of Top-1 accuracy; HRank yielded a compression ratio of 68.24\% and an accuracy loss of 3.01\%. These results proved that as compared to the current state of arts, our pruning algorithm can achieve higher parameter compression rate but with less network performance loss.

Table \ref{table 3} shows the network compression results for ILSVR-2012. For a global variance contribution rate of 99\%, the Top-1 accuracy loss of our methods for the three networks: VGG16-bn(pruned\_0.99), VGG19-bn(pruned\_0.99), and ResNet50(pruned\_0.99) was between $1.9\% to 2.2\%$, which was bigger than the loss in Table \ref{table 1}. This performance loss difference was mainly caused by the task difficulty difference between CIFAR-10 and ILSVRC-2012 as the latter dataset has more categories than the former and is more difficult to be accurately classified. 

Table \ref{table 4} lists the method comparison results. The ones labeled by ``ours" were those pruned using the new pruning method proposed in this paper. Three other methods were compared: the feature reconstruction-based methods(ThiNet \cite{luo2017thinet}, NISP-50-B \cite{yu2018nisp} and GAL-1\cite{Lin2019Generative}), the Taylor-expansion-based method(Taylor-FO-BN\cite{Molchanov2019Taylor})and the information-theory-based method (HRank \cite{Lin2019HRank}). Using our hierarchical pruning method for a global variance contribution ratio of 0.99 (pruned\_0.99, ours), we obtained a parameter compression rate of 52.76\% with a Top-1 accuracy of 73.01\% and a Top-5 accuracy of 91.11\%. All other assessed methods yielded lower parameter compression rate but with lower Top-1 and Top-5 accuracy than our method although FLOPs were similar.


\section{Discussion and Conclusion}
We proposed a new method to estimate the network redundancy and a new hierarchical algorithm to efficiently and effectively prune network filters. The method for estimating network redundancy is inspired by the Hessian matrix degeneration analysis proposed by Orhan and Pitkow\cite{orhan2018skip} but we made a new contribution by estimating the redundancy or the maximum pruning ratio through the weight gradient matrix PCA. We also proposed a gradient preprocessing method to reduce the random error of parameter gradient. The hierarchical pruning algorithm was designed to condense the network both along the network hierarchy (layer-wise) and within each layer (filter-wise), providing an effective way to avoid over-pruned layers. As compared to current state-of-art filter-wise pruning approaches which can be treated as an ablation study of our methods, our pruning algorithm achieved a high pruning ratio and maintained a small network prediction accuracy loss.

The major novelty of this work was that we provided a comprehensive solution for the three standing problems in network compression. Gradient matrix analysis itself is not new but the use of it for estimating the maximum network compression ratio is new. The hierarchical pruning method is not a simple extension of current filter-wise pruning but a result of careful consideration of both efficiency and robustness and represents a first-of-its-kind method to the best of our knowledge. CE has been used in previous network compression research \cite{Dong2017Surgeon,bao2018cross}. Our methods differ from the previous ones by measuring CE from the weight distribution of adjacent layers. CE in Bao et al was calculated from the network output before and after deleting a weight which is totally different from our approach. The methods in \cite{Dong2017Surgeon,bao2018cross} are still filter or weight-wise methods. Ours is a layer-first-filter-second hierarchical method, which is totally different from the previous methods including the above two.  
{\small
\bibliographystyle{IEEEtran}

}

\begin{IEEEbiography}[{\includegraphics[width=1in,height=1.25in,clip,keepaspectratio]{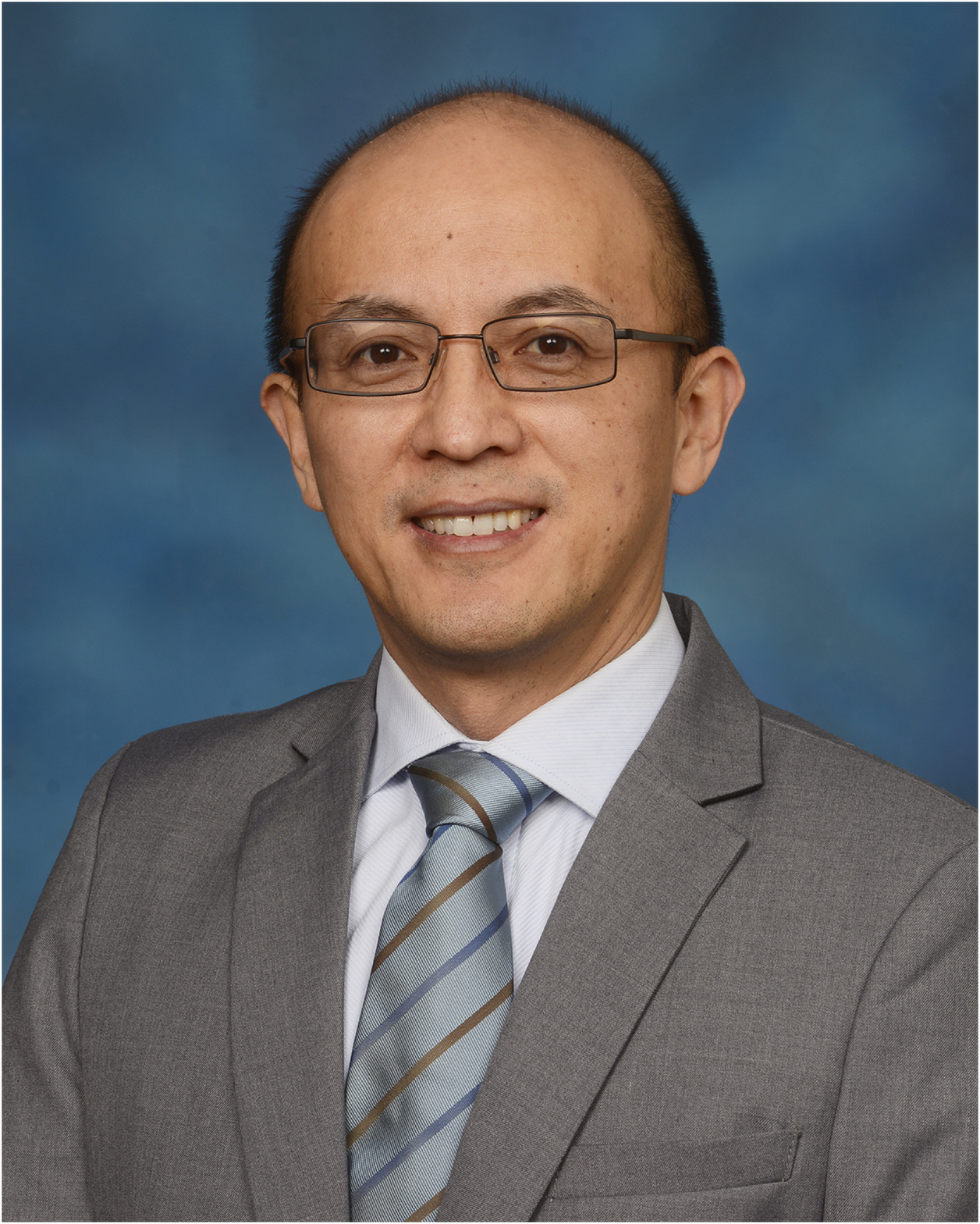}}]{Ze Wang}
is currently an Associate Professor with the Department of Diagnostic Radiology and Nuclear Medicine, University of Maryland School of Medicine, Baltimore, MD.
\end{IEEEbiography}

\begin{IEEEbiography}[{\includegraphics[width=1in,height=1.25in,clip,keepaspectratio]{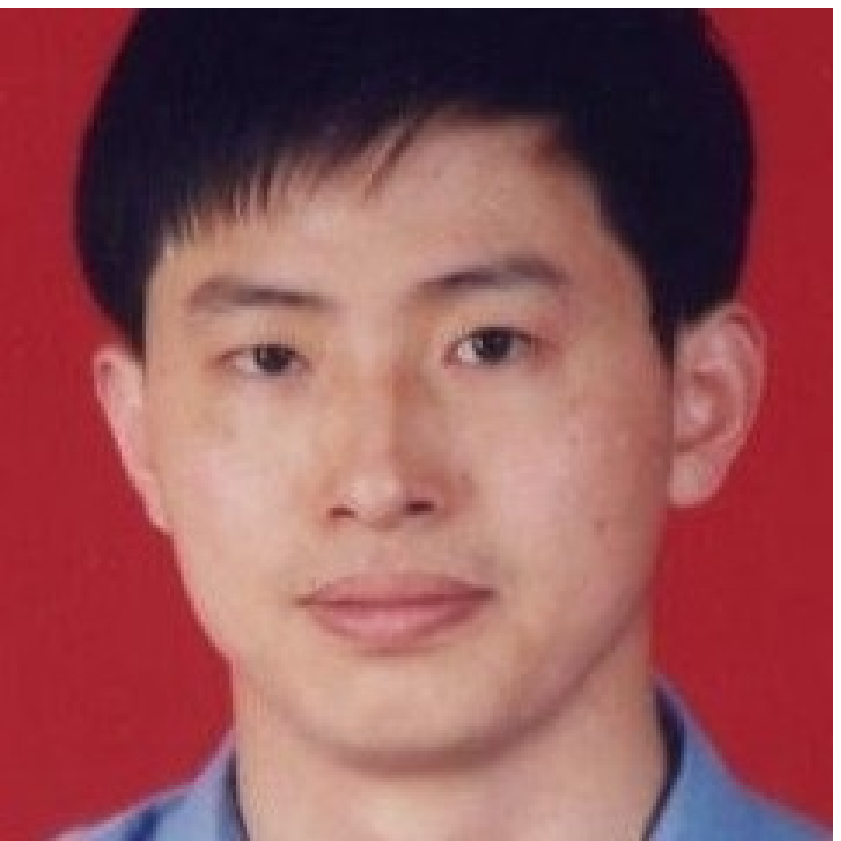}}]{Yilong Yin} is currently a Professor with the Research Center of Artificial Intelligence, Shandong University and School of Software, Jinan, China. His research interests include machine learning and data mining, medical image processing and biological recognition.
\end{IEEEbiography}

\begin{IEEEbiography}[{\includegraphics[width=1in,height=1.25in,clip,keepaspectratio]{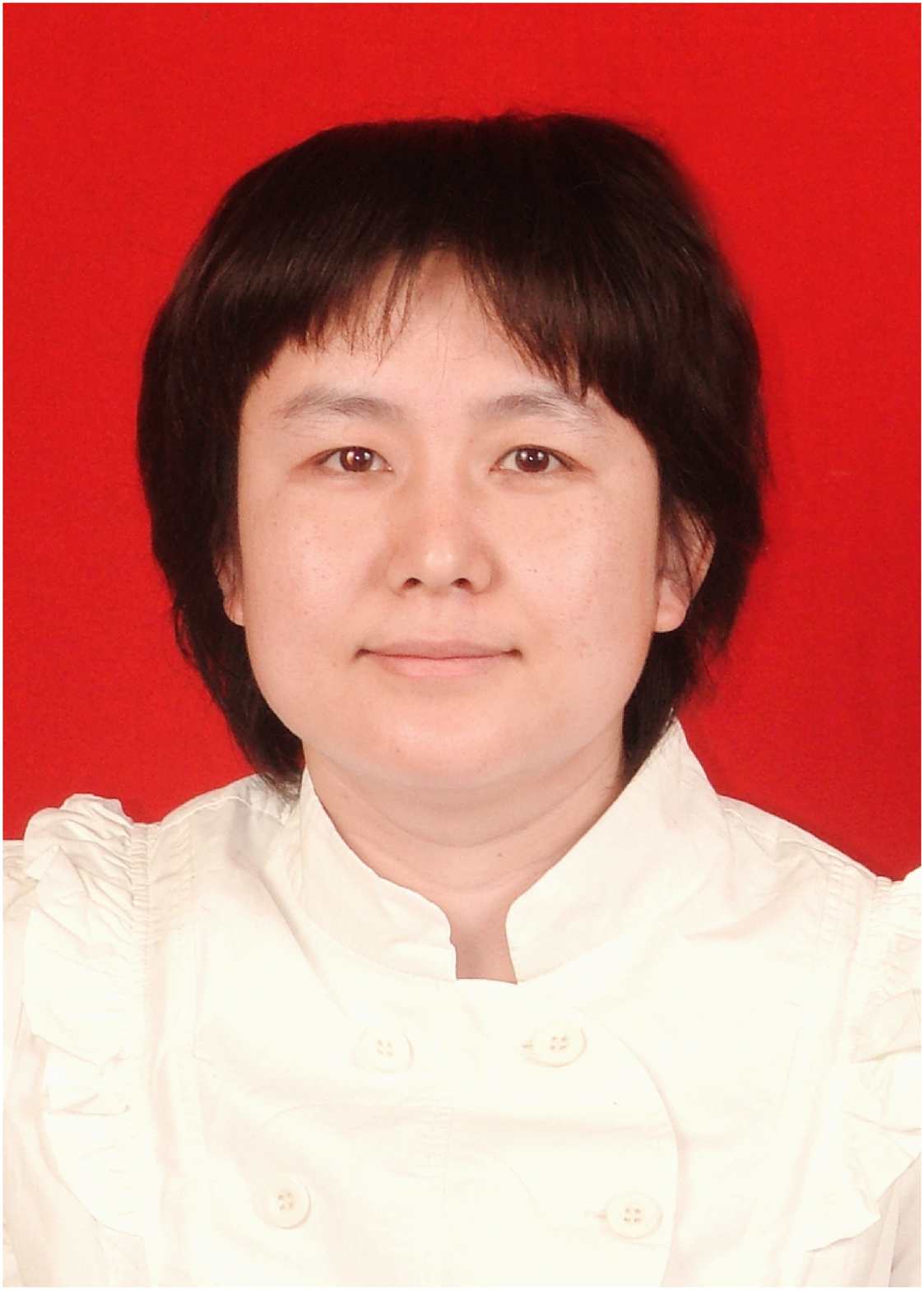}}]{Li Lian} is currently an  Associate Professor with the Research Center of Artificial Intelligence, Shandong University and School of Software, Jinan, China. Her research interests include knowledge representation and reasoning in information retrieval \& social networks.
\end{IEEEbiography}

\begin{IEEEbiography}[{\includegraphics[width=1in,height=1.25in,clip,keepaspectratio]{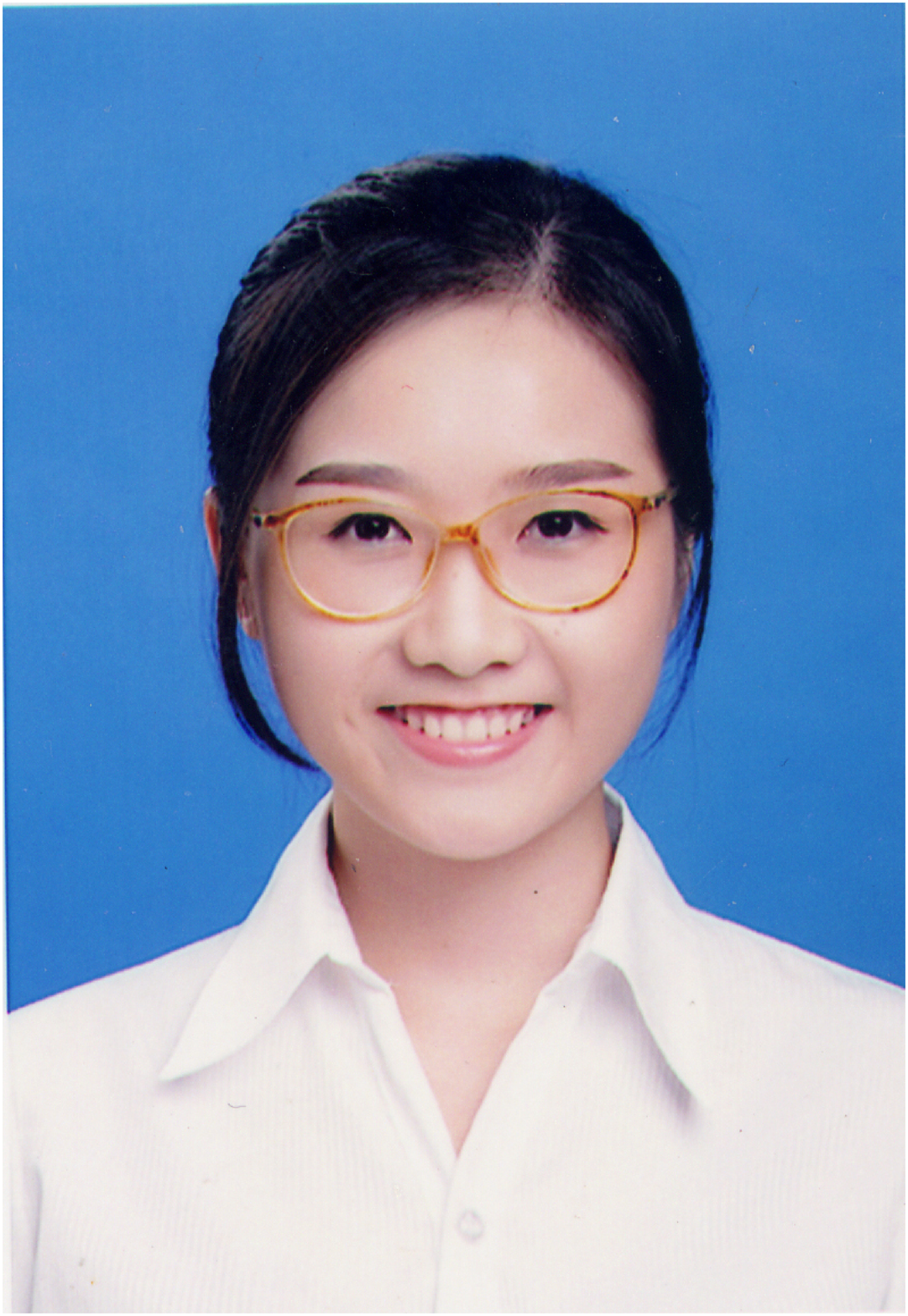}}]{Ziqi Zhou}
is currently working toward the graduate degree at the school of software, Shandong University, Jinan, China. Her research interests focus on machine learning.
\end{IEEEbiography}

\end{document}